\documentclass{article}
\usepackage{spconf,amsmath,graphicx}
\usepackage{multirow}
\usepackage{amssymb}
\usepackage{amsfonts}
\usepackage{booktabs}

\title{Precognition in Task-Oriented Dialogue Understanding: Posterior Regularization by Future Context}
\name{Nan Su, Yuchi Zhang, Chao Liu, Bingzhu Du, Yongliang Wang
\thanks{Contacted email: sujiu.sn@antgroup.com}\thanks{*: Financial NLU will be released recently, please continue to follow us.}}
\address{Ant Group, Beijing, China}
\begin{document}
\ninept
\maketitle
%

\begin{abstract}
Task-oriented dialogue systems have become overwhelmingly popular in recent researches. Dialogue understanding is widely used to comprehend users' intent/emotion/dialogue state in task-oriented dialogue systems. Most previous works on such discriminative tasks only models current query or historical conversations. Even if in some work the entire dialogue flow was modeled, it is not suitable for the real-world task-oriented conversations as the future contexts are not visible in such cases. In this paper, we propose to jointly model historical and future information through the posterior regularization method. More specifically, by modeling the current utterance and past contexts as prior, and the entire dialogue flow as posterior, we optimize the KL distance between these distributions to regularize our model during training. And only historical information is used for inference. Extensive experiments on two dialogue datasets validate the effectiveness of our proposed method, achieving superior results compared with all baseline models.

\end{abstract}
%

\section{Introduction}
\label{sec:intro}

Building a task-oriented dialogue system that can interact with humans naturally is a challenging yet interesting research in the field of artificial intelligence. This field has received increasing attention recently, due to its wide range of applications, such as XiaoIce, AliMe and Zhixiaobao\cite{zhou2020design,li2017alime,lou2020efficiently,xuan2021sead}. The goal of this type of task-based dialogue system is to help the user achieve a purpose, such as hotel reservation or receiving investment suggestions. 
Dialogue understanding plays a vital role in such dialogue systems. For instance, a typical task of dialogue understanding is to identify the user's intents\cite{hristidis2018chatbot} by capturing the semantics of the current utterance and the context throughout the conversation, which can be regarded as a classification problem for each utterance \cite{gupta2019casa}. Emotion recognition is another example of dialogue understanding and is usually used to identify user's emotion in a dialogue flow.

\begin{figure}[h]
	\centering  
	\includegraphics[width=0.85\linewidth]{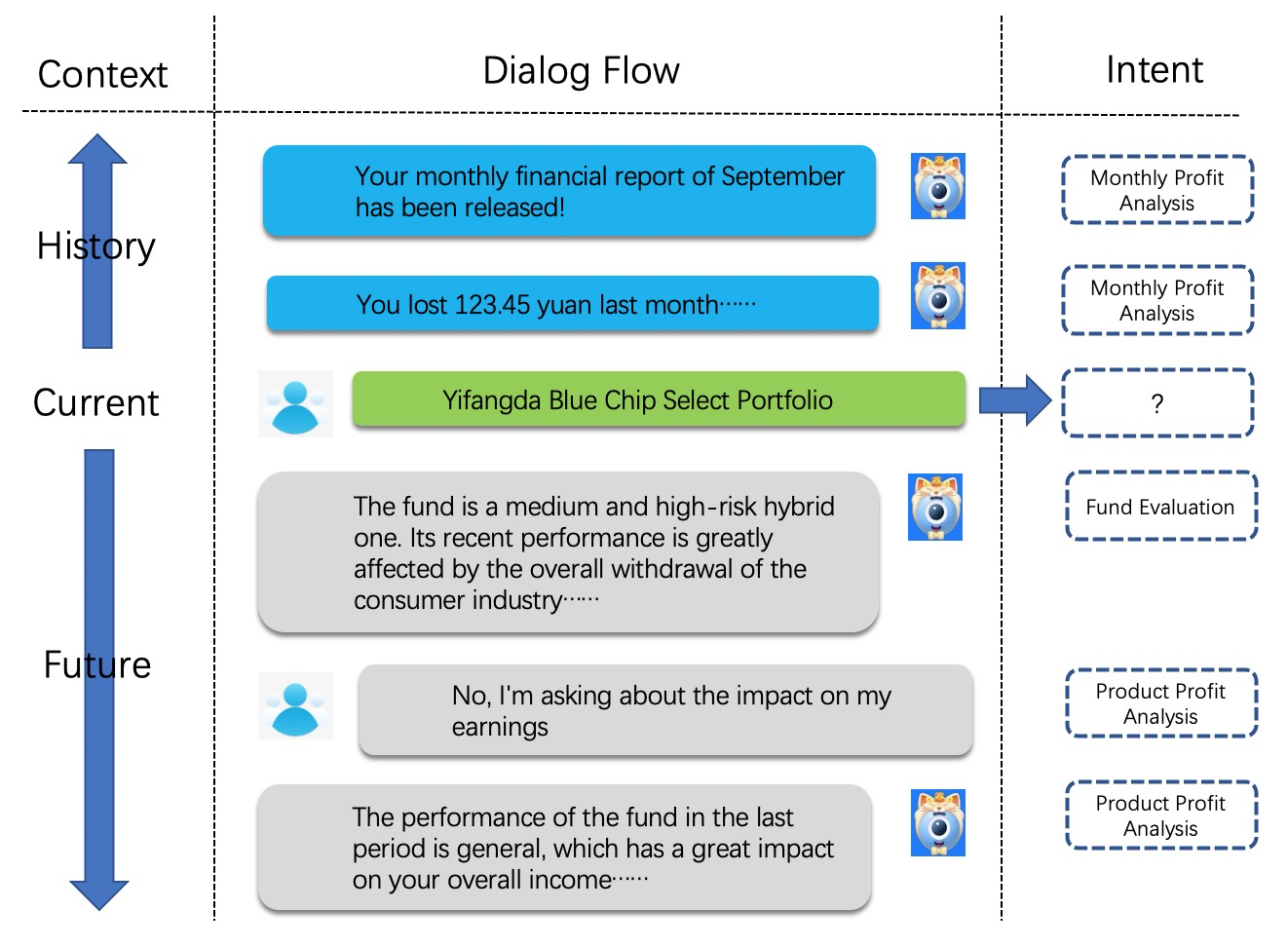} 
	\vspace{-0.2cm}
	\caption{Snippet of a sample conversation in Zhixiaobao with intents shown in dashed boxes}  
	\vspace{-0.4cm}
	\label{fig:demo4zxb}   
\end{figure}

Nowadays, human-computer interaction has become more complex and relies more on dialogue understanding techniques\cite{huang2018flowqa,gupta2020conversational}. 
Traditional dialogue systems are modeled causally, which means that dialogue systems can only obtain the current dialogue state from historical information and cannot go beyond the information of future dialogues. However the future dialogues contains additional information that should not be overlooked.\cite{shen2018nexus,fu2020context}

Figure \ref{fig:demo4zxb} is a demonstration of a multi-turn conversation in Zhixiaobao (a financial virtual assistant). Given a query mentioning a single fund name, it is hard to predict user's intent even if we take into account the historical context (blue box). The intention is easier to predict when we look a few rounds ahead, as the user makes clear in the ``future" context (gray box) that his intention is to make a fund 
evaluation.


Shen et al.\cite{shen2018nexus} and Feng et al.\cite{feng2020regularizing} adopt variational inference techniques to address the future context modeling problem. By introducing a Gaussian distributed prior and optimizing the KL distance between the prior and the true posterior in the training phase, one can extract an implicit variable $z$ from the prior as the embedding vector for future contexts. However, this kind of work only focuses on natural language generation tasks, and there is a lack of relevant research in the field of 
dialogue understanding. Inspired by the work on ``twin networks"\cite{serdyuk2017twin}, we propose a posterior regularization method to jointly model historical and future information. By modeling the ``current + historical" context as the prior and the ``future + current + historical" context as the posterior, we optimize the KL distance between the two Multinomial distributions and use the prior distribution only for inference. We hypothesize that our approach solves the problem of modeling future contexts by implicitly forcing the ``current + historical" parameters to hold information about future contexts.


Our contributions are:
(1) We propose a posterior regularization method to jointly modeling historical information and future information in multi-turn dialog system.
(2) A large-scale multi-turn dialog dataset named ``Financial NLU"\textsuperscript{*} is built by collecting real world data of an online service called Zhixiaobao. 
(3) Experimental results on two dialogue datasets show that our approach outperforms the baseline dialogue understanding approach, which demonstrates that leveraging future dialogues outside the dialogue history through posterior regularization can improve the performance of contextual dialogue comprehension in terms of accuracy and F1 scores.

\section{Preliminaries}
\label{sec:preliminaries}
\subsection{Evidence Lower Bound (ELBO)}
Amortized variational inference (AVI) is a class of methods which approximate latent variable distributions $p$\cite{rezende2014stochastic,mnih2014neural,kingma2014autoencoding}. To achieve such approximation, a family of parameterized distributions $q(\mathbf{z})$, named as variational distributions, are introduced. KL divergence between such family of variational distributions and the true latent distribution ($D_{\text{KL}}(q\vert\vert p)$) is minimized. And thus real-world complex distributions $p$ can be similarly and conveniently modeled by these variational distributions $q$.

Variational Auto Encoder (VAE) \cite{kingma2014autoencoding}, an example of AVI, minimizes the divergence $D_{(\text{KL})}(q(\mathbf{z\vert \mathbf{x}}) \vert \vert p(\mathbf{z}\vert \mathbf{x}))$. For tractability, Evidence Lower Bound (ELBO)\cite{kingma2014autoencoding} is derived and optimized during training. In fact, in a generative model, ELBO is the lower bound of the logarithm likelihood $\log{p(\mathbf{x})}$ of data $\mathbf{x}$. Thus the likelihood of the data distribution is optimized while ELBO is optimized, and thus VAE-like models achieve high success by mapping image, audio, text etc. into latent space and generating such kinds of data by sampling from the modeled latent distributions\cite{bowman-etal-2016-generating}.

For a discriminative task, however, things are quite different. Instead of modeling the distributions of the original data, we focus on obtain the correct discrete category $y$, given the input data or feature $\mathbf{x}$. Hence $\log{p(y\vert \mathbf{x})}$ is optimized instead of $\log{p(\mathbf{x})}$. From a approach which is similar to but different from that in \cite{kingma2014autoencoding}, we can obtain the ELBO of a disciminative likelihood as follows:
\begin{equation}
\label{eq:elbo}
    \begin{split}
        \log{p(y\vert \mathbf{x})}&=\log{\int_{\mathbf{z}} p(y, \mathbf{z}\vert \mathbf{x})} \\
        &=\log{\int_{\mathbf{z}}p(y\vert \mathbf{z},\mathbf{x})p(\mathbf{z}\vert\mathbf{x})} \\
        &=\log{\int_{\mathbf{z}}\frac{p(y\vert \mathbf{z},\mathbf{x})p(\mathbf{z}\vert\mathbf{x})q(\mathbf{z})}{q(\mathbf{z})}} \\
        &=\log{\mathbb{E}_{q(\mathbf{z})}}[p(y\vert \mathbf{z},\mathbf{x})\frac{p(\mathbf{z}\vert \mathbf{x})}{q(\mathbf{\mathbf{z}})}] \\
        &\geq \mathbb{E}_{q(\mathbf{z})}[\log{(p(y\vert\mathbf{z},\mathbf{x})\frac{p(z\vert \mathbf{x}))}{q(\mathbf{z})})}] \\
        &=\mathbb{E}_{q(\mathbf{z})}[\log{p(y\vert \mathbf{x},\mathbf{z})}-\log{\frac{q(\mathbf{z})}{p(\mathbf{z}\vert \mathbf{x})}}] \\
        &=\mathbb{E}_{q(\mathbf{z})}[\log{p(y\vert \mathbf{x},\mathbf{z})}]-D_{\text{KL}}(q(\mathbf{z})\vert\vert p(\mathbf{z}\vert \mathbf{x}))
    \end{split}
\end{equation}
where $\mathbf{z}$ is the vector of the latent space, which can be regarded as the representation of the input data.

The first term of Equation.\ref{eq:elbo} is a classification objective, taking $\mathbf{x}$ and its representation $\mathbf{z}$ into account. And the second term is the regularization term which minimizes the divergence between variational disritbution and the latent representation distribution.

\begin{figure*}[h]
\renewcommand{\baselinestretch}{0.9}
	\centering  
	\includegraphics[width=0.7\linewidth]{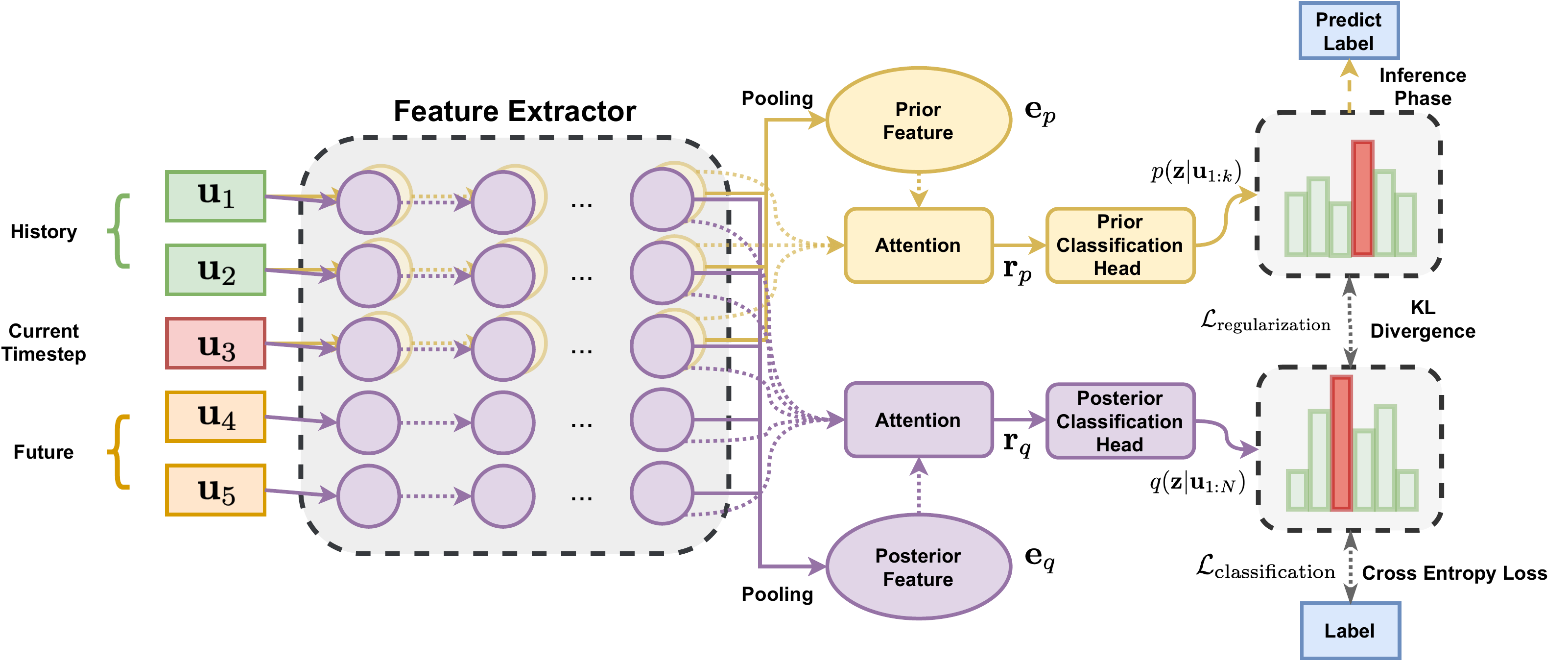}  
	\vspace{-0.2cm}
	\caption{Overview of the model framework of the proposed method. A dialogue session of 5 utterances in total is shown for instance, and it is currently at the 3rd turn. Prior (yellow) backbone extracts feature of dialogue histories and the current query, while posterior (purple) backbone extracts feature of the whole dialogue. $\mathbf{r}_p$ and $\mathbf{r}_q$ are derived through pooling and attention mechanism. Then they are passed through two classification heads to get $p(\mathbf{z}\vert \mathbf{u}_{1:k})$ and $q(\mathbf{z}\vert \mathbf{u}_{1:N})$. In the training phase, $q$ is used as the final distribution for $\mathcal{L}_{\text{classification}}$ calculation. While in the inference phase, we use $p$ for classification. Regularization objective $\mathcal{L}_{\text{regularization}}$ between $q$ and $p$ is also optimized during training. Such regularization and the posterior part only exist in the training phase, thus causality is not violated.} 
	\vspace{-0.4cm}
	\label{fig:posterior}   
\end{figure*}

\subsection{Posterior Regularization}
In discriminative situations, there's no problem to calculate and optimize the second term in Equation.\ref{eq:elbo}. Nevertheless, problem occurs when we come to the first one. $\mathbb{E}_{q}[\cdot]$ implies that a sampling process exists during the calculation of the classification likelihood. Unlike a generative task which samples $\mathbf{z}$ from a continuous latent space\cite{bowman-etal-2016-generating}, sampling from a discrete distribution introduces extra and big variance for feature extracting. And since the nature of classification is to extract accurate and stable representations, variances in feature extracting process by sampling might harm the performance of the classification results. Moreover, labels in classification tasks are one-hot and never follow multi-peak distributions. Therefore, for discriminative tasks we adopt point estimation rather than sampling in the first term of Equation.\ref{eq:elbo}.
It is also proved in \cite{deng2018latent} that $|\mathbb{E}_{q(\mathbf{z})}[f(\mathbf{x},\mathbf{z})]-f(\mathbf{x},\mathbb{E}_{q(\mathbf{z})}[\mathbf{z}])|$\par\noindent$ \leq C$, where $C$ is the upper bound of the norm of the Hessian matrix with respect to $\mathbf{z}$, i.e.  $||H_{f}(\mathbf{z})||_{F}$ where $F$ is the Frobenius norm \cite{deng2018latent,bishop1992exact,bottcher2008frobenius}. In a classification problem \cite{deng2018latent} has pointed out that the result of $f$ and $\mathbf{z}$ are both simplex and hence $C$ is small, thus such term with a sampling process can be substituted by a deterministic function. For a discriminative task $f$ is $p(y\vert\mathbf{x},\mathbf{z})$ in detail, therefore the final lower bound is:
\begin{equation}
    \label{eq:disc}
    \begin{split}
        \mathcal{L}_{\text{ELBO}}&=\log{p(y\vert \mathbf{x},\mathbb{E}_{q(\mathbf{z})}[\mathbf{z}])}-D_{\text{KL}}(q(\mathbf{z})\vert\vert p(\mathbf{z}\vert \mathbf{x})) \\
        &=\log{p(y\vert \mathbf{x},g(\mathbf{x}))}-D_{\text{KL}}(q(\mathbf{z})\vert\vert p(\mathbf{z}\vert \mathbf{x}))
    \end{split}
\end{equation}
where $\mathbf{z}$ is no longer sampled from $q$, but is mapping to a representation directly by $g$ which directly predicts its mean value, taken $\mathbf{x}$ as input. In our dialogue understanding problem, $y$ is the intent/emotion/state, and current query and dialogue histories as $\mathbf{x}$.

Experiments also show that sampling method perform poorly in our dialogue understanding task. We'll discuss this in section \ref{sec:experiments}. 

Equation.\ref{eq:disc} now consists of an ordinary classification likelihood as the first term. And the second term regularize latent distribution of $\mathbf{z}$ by $q(\mathbf{z})$. We set the posterior distribution $q$ the function of both the dialogue history and the future, therefore such method is named as posterior regularization.

\section{Methodology}
Consider a dialogue session $[\mathbf{u}_1,\mathbf{u}_2,...,\mathbf{u}_N]$, where $\mathbf{u_i}=[u_{i1},u_{i2},$\par\noindent$...,u_{in_{i}}]^T$ is the $i_{th}$ utterance, $u_{ij}$ is the $j_{th}$ word of the $i_{th}$ utterance, and $n_{i}$ is the length of the $i_{th}$ utterance. Suppose the current utterance is $\mathbf{u}_{k}$, i.e. we are now at the $k_{th}$ turn of the conversation. Dialogue understanding task is a classification task which aims to find the correct intent, emotion, tagging, etc. of query $\mathbf{u}_{k}$, given the query itself and all dialogue histories $\mathbf{u}_{<k}$.

As we mentioned in section \ref{sec:intro}, a real-world dialogue flow is a casual sequence. Therefore, existing methods only make use of the past sequences $[\mathbf{u}_1,...,\mathbf{u}_{k}]$ for such classification task, both in training and inferring phases. Future part of the contexts $[\mathbf{u_{k+1},...,\mathbf{u}_{N}}]$ contributes nothing to the result of the $k_{th}$ turn, since consistency and causality should not be violated.

We managed to solve the above dilemma by leveraging future utterances by posterior regularization. Fig.\ref{fig:posterior} shows the whole framework of our method. Our model consists of posterior units which are shown in purple and prior units which are shown in yellow. The prior part models the current query along with dialogue histories, which is the same as baseline methods. And the posterior part owns a ``future sight", which extracts the semantic information of the whole dialogue session, including the future part in addition to that of the prior.

\subsection{Feature Extracting}
In the primary stage, the feature of both the prior and posterior inputs are extracted. Firstly, utterance sequences are flattened into word level sequences as inputs. For the prior part, dialogue history and the current query $[\mathbf{u}_1,...,\mathbf{u}_k]$ are flattened into $I_{p}=[u_{11},u_{12},...,u_{k1},u_{k2},...,u_{kn_{k}}]$. And for the posterior part, the whole session $[\mathbf{u}_1,...,\mathbf{u}_k,...,\mathbf{u}_N]$ is flattened into $I_{q}=[u_{11},u_{12},...,u_{kn_{k},...,u_{Nn_{N}}}]$.

Then the intermediate representation of both the prior and posterior parts are obtained:
\begin{gather}
    \mathbf{e}_{p} = \text{Pooling}[F_p(I_p)] = \text{Pooling}[E_p] \\
    \mathbf{e}_{q} = \text{Pooling}[F_q(I_q)] = \text{Pooling}[E_q]
\end{gather}
where $F_p$ and $F_q$ are feature extractor backbones, usually pretrained transformer models, we adopt Bert\cite{jocab2018bert} in our experiments. As a result, $E_p=[\mathbf{e}_{11},...,\mathbf{e}_{kn_{k}}]$ and $E_q=[\mathbf{e}_{11},...,\mathbf{e}_{Nn_{N}}]$ are extracted contextual embeddings of $I_p$ and $I_q$. $\text{Pooling}$ methods reduce $E_p$ and $E_q$ into single vector representations $\mathbf{e}_{p}$ and $\mathbf{e}_{q}$ respectively, and we adopt average pooing, or just use the embedding of the ``[CLS]" step\cite{jocab2018bert} in practice.

Attention\cite{vaswani2017attention} between the pooling representations and contextual embeddings are used to obtain the final prior and posterior representation vectors:
\begin{gather}
    \mathbf{r}_{p} = \text{Attn}(\mathbf{e}_p, E_p) \label{eq:representation_p} \\
    \mathbf{r}_{q} = \text{Attn}(\mathbf{e}_q, E_q) \label{eq:representation_q}
\end{gather}
where $\text{Attn}(\mathbf{e}, E)$ means attention between query vector $\mathbf{e}$ and a series of key and value $E$, the details of calculation is the same as that in \cite{vaswani2017attention}.

\subsection{Classification and Distribution Regularization}
In the second stage, we obtain both the prior and posterior classification probabilities. Given the representations $\mathbf{r}_p$ and $\mathbf{r}_q$ derived from \ref{eq:representation_p} and \ref{eq:representation_q} respectively, prior and posterior are modeled as multinomial distributions:
\begin{gather}
    p(\mathbf{z}\vert \mathbf{u}_{1:k}) = \text{Softmax}(\text{MLP}_{\theta}(\mathbf{r}_p)) \label{eq:prior} \\
    q(\mathbf{z}\vert \mathbf{u}_{1:N}) = \text{Softmax}(\text{MLP}_{\varphi}(\mathbf{r}_q))
\end{gather} \label{eq:posterior}
where $\text{MLP}_{\theta}$ and $\text{MLP}_{\varphi}$ are multi-layer fully connected neural networks, and the prior and posterior categorical distributions are denoted by $p$ and $q$ respectively.

According to the discussion in section \ref{sec:preliminaries}, unlike generating tasks, we do not sample from the two distributions to obtain the final categorical distribution for classification. Needless to say, we simply adopt the distributions in \ref{eq:prior} and \ref{eq:posterior}: In the training phase, we select $q$ as the final likelihood, i.e. $p(y\vert \mathbf{u}_{<k})=\mathcal{I}[q(\mathbf{z}\vert \mathbf{u}_{1:N})]$ where $\mathcal{I}[\cdot]$ is the identical mapping. And since we cannot see the future in the inference phase, we choose $p(y\vert \mathbf{u}_{<k})=\mathcal{I}[p(\mathbf{z}\vert \mathbf{u}_{1:k})]$. As common classification tasks do, cross entropy applied are used to optimize such likelihood.

The most important part is the distribution regularization term, which is the same as the second term in Equation.\ref{eq:disc} as we discussed in section \ref{sec:preliminaries}. Thus the KL divergence is adopted to regularize $p(\mathbf{z}\vert \mathbf{u}_{1:k}) = \text{Softmax}(\text{MLP}_{\theta}(\mathbf{r}_p))$ by $q(\mathbf{z}\vert \mathbf{u}_{1:N}) = \text{Softmax}(\text{MLP}_{\varphi}(\mathbf{r}_q))$. Therefore, we obtain the final objective function of our method as follows:
\begin{equation}
    \begin{split}
        \mathcal{L}&=\mathcal{L}_{\text{classification}}+\lambda \mathcal{L}_{\text{regularization}} \\
        &=\log{p(y\vert \mathbf{u}_{\leq k}, \mathbf{r})}-\lambda D_{\text{KL}}(q(\mathbf{z}\vert \mathbf{u}_{1:N})\vert\vert p(\mathbf{z}\vert \mathbf{u}_{1:k}))
    \end{split}
\end{equation}
where $\lambda$ is a coefficient for KL annealing\cite{bowman-etal-2016-generating} in $[0,1]$. And representation $\mathbf{r}=\mathbf{r}_q$ in the training phase, while $\mathbf{r}=\mathbf{r}_p$ in the inference phase.



\section{Experiments}
\label{sec:experiments}
\subsection{Experimental Setups}
Our experiments are conducted on two conversational datasets: 
i)The emotion recognition dataset \textbf{IEMOCAP}\cite{Busso2008} contains videos of two-way conversations of ten unique speakers.
\begin{table}[htb]
\centering
\setlength{\tabcolsep}{5.5mm}
\begin{tabular}{c|c|c}
\toprule[1.5pt]
Dataset                  & Partition & \.Examples \\ \hline
\multirow{2}{*}{IEMOCAP} & train+val & 5810       \\ \cline{2-3} 
                         & test      & 1623       \\ \hline
\multirow{2}{*}{FCNLU}   & train+val & 8587       \\ \cline{2-3} 
                         & test      & 1082       \\ \bottomrule[1.5pt]
\end{tabular}
\caption{Training, validation and test data distribution in the datasets}
\label{tab:datasets}
\end{table}

\begin{table*}[htb]
\centering
\resizebox{.95\linewidth}{!}{
\begin{tabular}{c|cccccccc}
\toprule[1.5pt]
\multirow{2}{*}{Method}                         & acc                  & \multicolumn{1}{c|}{f1}                   & acc                  & \multicolumn{1}{c|}{f1}                   & acc                  & \multicolumn{1}{c|}{f1}                   & acc                  & f1     
\\ \cline{2-9}
                        & \multicolumn{2}{c|}{ws=10}                                       & \multicolumn{2}{c|}{ws=8}                                        & \multicolumn{2}{c|}{ws=6}                                        & \multicolumn{2}{c}{ws=4}            
              \\ \hline
TH-PH                   & 57.78$\vert$53.91          & \multicolumn{1}{c|}{57.73$\vert$53.79}          & 57.73$\vert$53.73          & \multicolumn{1}{c|}{57.60$\vert$53.78}          & 57.86$\vert$52.99          & \multicolumn{1}{c|}{57.73$\vert$53.01}          & 57.61$\vert$52.87          & 57.50$\vert$52.73          \\
TW-PW                   & 59.03$\vert$54.96          & \multicolumn{1}{c|}{59.00$\vert$54.79}          & 59.09$\vert$54.90          & \multicolumn{1}{c|}{58.90$\vert$54.72}          & 58.84$\vert$54.96          & \multicolumn{1}{c|}{58.71$\vert$54.88}          & 58.41$\vert$55.63          & 58.25$\vert$55.55          \\ \hline
TW-PH-S & $23.48\vert14.36$  & \multicolumn{1}{c|}{$19.04\vert14.00$} & $19.72\vert14.17$ & \multicolumn{1}{c|}{$18.61\vert13.78$} & $20.27\vert13.86$ &
\multicolumn{1}{c|}{$19.00\vert13.44$} & $18.79\vert12.82$ &
\multicolumn{1}{c}{$17.75\vert12.25$}
\\
\textbf{TW-PH}          & \textbf{58.90}$\vert$\textbf{54.53} & \multicolumn{1}{c|}{\textbf{58.71}$\vert$\textbf{54.55}} & \textbf{58.78}$\vert$\textbf{54.78} & \multicolumn{1}{c|}{\textbf{58.70}$\vert$\textbf{54.72}} & \textbf{58.29}$\vert$\textbf{54.34} & \multicolumn{1}{c|}{\textbf{58.16}$\vert$\textbf{54.27}} & \textbf{57.92}$\vert$\textbf{55.51} & \textbf{57.89}$\vert$\textbf{55.38} \\ \bottomrule[1.5pt]
\end{tabular}
}
\caption{Results on IEMOCAP dataset. ws represents window size of past contexts. Left: results with bi-GRU structure, right: results without bi-GRU structure.}
\label{tab:iemocap_results}
\end{table*}

Each video contains a single dyadic dialogue, segmented into utterances. Each utterance of a dialogue is annotated
with one of six emotion labels, which are
happy, sad, neutral, angry, excited, and frustrated. We use each utterance with its past and future contexts in the same conversation to form a training/validation/test example. 
ii) 
The Financial-domain Contextual Natural Language Understanding(\textbf{FCNLU}) dataset consists of multi-turn dialogues between a user and a task-oriented chatbot, collected from an online service. In each conversation, the user's utterance in the middle position is annotated with 97 intent types (e.g. fund evaluation, monthly profit analysis etc.) to ensure it has both past and future contexts to experiment with.

We conduct three types of experiments on both datasets and one extra type of experiment on IEMOCAP dataset as following:

\noindent \textbf{TH-PH} training and inference are performed with only the current utterance and its dialogue histories, which is implemented to form a baseline.

\noindent \textbf{TW-PW} training and inference are performed with the entire dialogue. We implement this experiment to demonstrate how much impact the futures can have on our tasks. Several sizes of future context windows are attempted to find the best reference. 

\noindent \textbf{TW-PH} training is performed with the whole conversation using the Posterior Regularization method. Inference is performed with only the histories.

\noindent \textbf{TW-PH-S}(IEMOCAP only) training is performed with the whole conversation using the sampling method, and inference is performed with only the histories. We implement this experiment to demonstrate that the point estimation performs better than sampling method for discriminative tasks, which has been discussed in previous sessions.



We use the Bert\cite{jocab2018bert} model pre-trained on large-scale financial conversation data as context encoder for the FCNLU dataset. The model has 12 hidden layers, each with 768 units and 12 self-attentive heads. The dimension of the hidden states used for attention is 100. During training we use dropout \cite{nitish2014dropout} on the BERT hidden units with a rate of 30\% to prevent overfitting. For optimization we use Adam optimizer \cite{kingma2015adam}. The initial learning rate is set to $5e^{-5}$. We conduct training with a multi-step LR, which decays the learning rate by 0.5 once the number of epoch reaches 2,4,6 and 8. 

For the IEMOCAP dataset, a bi-directional RNNs with Gated Recurrent Units(GRU)\cite{chung2014empirical} model is used to generate the context-aware representations for all utterances. The hidden dimension of RNN is set to 712, which is consistent with the feature dimension. The hidden dimension of the attention part is 128. For optimization an Adam optimizer is adopted and the learning rate is set to $1e^{-4}$.

We use KL-annealing(KLA) in all TW-PH experiments for both datasets. Specifically, the KL loss is added only after the model has been trained for 2 epochs on FCNLU dataset and 5 epochs on IEMOCAP dataset. After that, the weight of the KL loss is gradually increased from 0 to 1.
\begin{table}[!htb]
\centering
\setlength{\tabcolsep}{5.5mm}
\begin{tabular}{c|ccc}
\toprule[1.5pt]
\multirow{2}{*}{Method} & \multicolumn{3}{c}{acc}                                                                   \\ \cline{2-4} 
                        & ws=5                                & ws=3                                & ws=1           \\ \hline
TH-PH                   & \multicolumn{1}{c|}{74.58}          & \multicolumn{1}{c|}{71.90}          & 71.16          \\
TW-PW                   & \multicolumn{1}{c|}{75.51}          & \multicolumn{1}{c|}{74.49}          & 73.57          \\ \hline
\textbf{TW-PH}          & \multicolumn{1}{c|}{\textbf{74.86}} & \multicolumn{1}{c|}{\textbf{74.03}} & \textbf{73.57} \\ \bottomrule[1.5pt]
\end{tabular}
\caption{Results on FCNLU dataset.  ws represents window size of past contexts.}
\label{tab:fcnlu_results}
\end{table}

\subsection{Results}

We compare the performance of our proposed methods with the baseline method for various past window size in Table \ref{tab:iemocap_results} and Table \ref{tab:fcnlu_results}. Our proposed method outperforms all baseline models on both datasets.

\noindent \textbf{IEMOCAP:} For the IEMOCAP dataset, our model surpasses the baseline approach by accuracy and f1-scores of up to 1.12\% and 1.10\%, respectively. We carry out experiments with different past context window sizes and find that our proposed method helps improve the performances in all cases. This confirms our hypothesis that the future information brought during training can lead to a better understanding of the current state of the conversation. Furthermore, as an ablation study, we remove the bi-GRU structure from the models and find that while the performance of all models generally decreased, the accuracy and f1 scores of our proposed method were steadily higher than those of the baseline model by about 0.8\%-1.5\%, depending on the different window sizes.
Besides, we verify our hypothesis that using sampling method in the discriminative task is not as effective as the generation task, since the model performance decays significantly in all cases, as shown in Table \ref{tab:iemocap_results}. 

\noindent \textbf{FCNLU:} As evidenced by Table \ref{tab:fcnlu_results}, for the FCNLU dataset, the model training with posterior regularization outperforms baseline model by 2.41\% accuracy in the best case. Although reducing the window size of past contexts leads to decrease in model performance, bringing future information into the training process actually compensates for this gap, resulting in accuracy scores of 74.86\%, 74.03\% and 73.57\%. We believe that the model has difficulty capturing enough information when past conversations are limited. The model performance can therefore be enhanced by 
learning from the whole dialogue flow. The less the model sees from the history, the more the model benefits from the future contexts through the posterior regularization method.

\section{Conclusion}

In this paper, we present a posterior regularization method to jointly modeling historical information and future information. Meanwhile, we built a multi-turn dialog database called 	``Financial NLU" which collected from mature internet financial platform, named	``Zhixiaobao", and it will be released recently. Experimental results validate the effectiveness of the proposed method, which offers promising results on IEMOCAP and FCNLU datasets.

\bibliographystyle{IEEEbib}
\bibliography{strings,refs}

\end{document}